\documentclass[10pt,twocolumn,letterpaper]{article}

\usepackage{iccv}
\usepackage[accsupp]{axessibility}

\usepackage{graphicx}
\usepackage{amsmath}
\usepackage{amssymb}
\usepackage{booktabs}
\usepackage{algorithm}
\usepackage{algpseudocode}
\usepackage{listings}
\usepackage{subcaption}

\usepackage[breaklinks=true,bookmarks=false]{hyperref}
\iccvfinalcopy 

\usepackage[capitalize]{cleveref}
\crefname{section}{Sec.}{Secs.}
\Crefname{section}{Section}{Sections}
\Crefname{table}{Table}{Tables}
\crefname{table}{Tab.}{Tabs.}

\ificcvfinal\fi

\newcommand{\website}{\href{https://agi-labs.github.io/manipulate-by-seeing/}{https://agi-labs.github.io/manipulate-by-seeing/}}

\usepackage{etoolbox}
\makeatletter
\AfterEndEnvironment{algorithm}{\let\@algcomment\relax}
\AtEndEnvironment{algorithm}{\kern2pt\hrule\relax\vskip3pt\@algcomment}
\let\@algcomment\relax
\newcommand\algcomment[1]{\def\@algcomment{\footnotesize#1}}
\renewcommand\fs@ruled{\def\@fs@cfont{\bfseries}\let\@fs@capt\floatc@ruled
  \def\@fs@pre{\hrule height.8pt depth0pt \kern2pt}%
  \def\@fs@post{}%
  \def\@fs@mid{\kern2pt\hrule\kern2pt}%
  \let\@fs@iftopcapt\iftrue}
\makeatother

\begin{document}

\title{Manipulate by Seeing: Creating Manipulation Controllers\\ from Pre-Trained Representations}

\author{ \hspace{-1.5em} Jianren Wang\thanks{Denotes equal contribution.} \hspace{1em} Sudeep Dasari$^*$ \hspace{1em} Mohan Kumar Srirama \hspace{1em} Shubham Tulsiani \hspace{1em} Abhinav Gupta \\ \\
Carnegie Mellon University
}

\maketitle

\begin{abstract}
The field of visual representation learning has seen explosive growth in the past years, but its benefits in robotics have been surprisingly limited so far. Prior work uses generic visual representations as a basis to learn (task-specific) robot action policies (e.g., via behavior cloning). While the visual representations do accelerate learning, they are primarily used to encode visual observations. Thus, action information has to be derived purely from robot data, which is expensive to collect! In this work, we present a scalable alternative where the visual representations can help directly infer robot actions. We observe that vision encoders express relationships between image observations as \textit{distances} (e.g., via embedding dot product) that could be used to efficiently plan robot behavior. We operationalize this insight and develop a simple algorithm for acquiring a distance function and dynamics predictor, by fine-tuning a pre-trained representation on human collected video sequences. The final method is able to substantially outperform traditional robot learning baselines (e.g., $70\%$ success v.s. $50\%$ for behavior cloning on pick-place) on a suite of diverse real-world manipulation tasks. It can also generalize to novel objects, without using  \textit{any} robot demonstrations during train time. For visualizations of the learned policies please check: \website.
\end{abstract}

\section{Introduction}
\label{sec:intro}

\begin{figure}[t]
    \centering 
    \includegraphics[width=\linewidth]{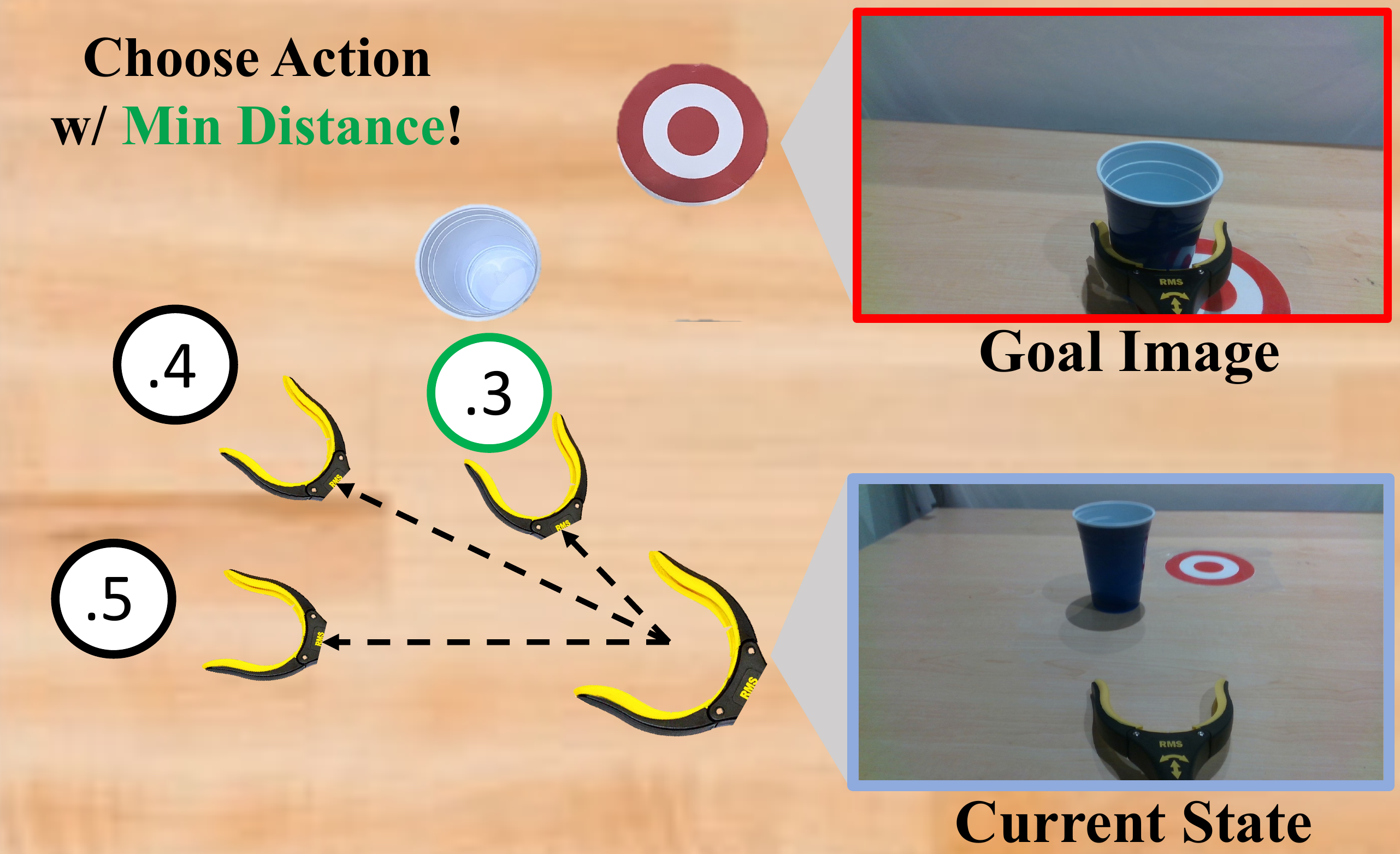} 
        \caption{This paper proposes to solve a range of manipulation tasks (e.g. pushing) by learning a functional distance metric within the embedding space of a pre-trained network. This distance function -- in combination with a learned dynamics model -- can be used to greedily plan for robot actions that reach a goal state. Our experiments reveal that the proposed method can outperform SOTA robot learning methods across four diverse manipulation tasks. }
    \label{fig:teaser}
    
\end{figure}

The lack of suitable, large-scale data-sets is a major bottleneck in robot learning. Due to the physical nature of data collection, robotics data-sets are: (a) hard to scale; (b) collected in sterile, non-realistic environments (e.g. robotics lab); (c) too homogeneous (e.g. toy objects with fixed backgrounds/lighting). In contrast, vision data-sets contain diverse tasks, objects, and settings (e.g. Ego4D~\cite{grauman2022ego4d}). Therefore, recent approaches have explored transferring priors from large scale vision data-sets to robotics settings. What is the right way to accomplish this?

Prior work uses vision data-sets to pre-train representations~\cite{parisi2022unsurprising,nair2022r3m,Radosavovic2022} that encode image observations as state vectors (i.e. $s = R(i)$). This visual representation is then simply used as an input for a controller learned from robot data -- e.g. a policy $\pi(a | s)$, trained with expert data (e.g., via Behavior Cloning~\cite{ross2011reduction}, Nearest Neighbors~\cite{pari2021surprising}, etc.), or a Value function $V(s)$, trained using exploratory roll-outs via Reinforcement Learning~\cite{sutton2018reinforcement}. Is this approach the most efficient way to use pre-trained representations? We argue that pre-trained networks can do more than just represent states, since their latent space already encodes semantic, task-level information -- e.g. by placing semantically similar states more closely together. Leveraging this structure to infer actions, could enable us to use significantly less robotic data during train time.

\begin{figure*}[t]
    \centering 
    \includegraphics[width=\linewidth]{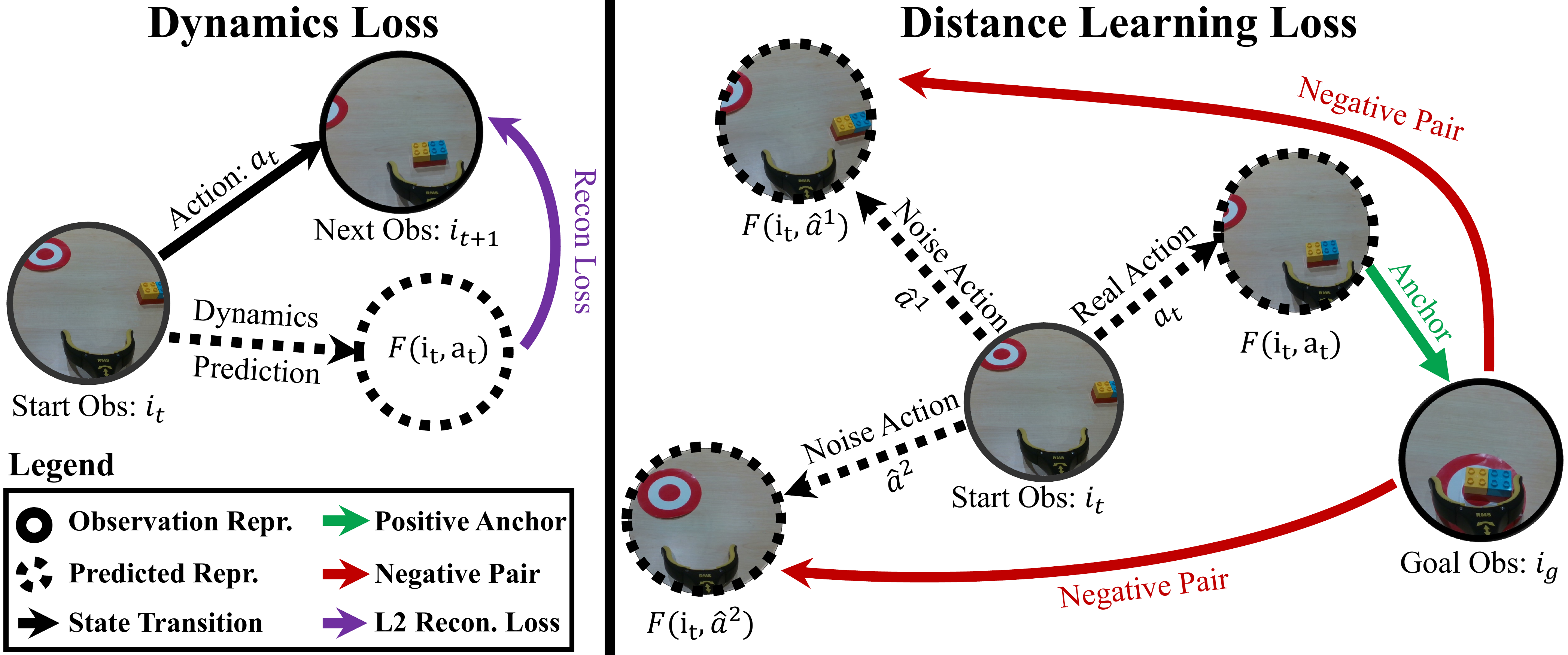} 
        \caption{We visualize the loss functions used to train our method. The dynamics function is trained via reconstruction loss in embedding space (left). The distance function is trained via contrastive learning, with positive anchors chosen by predicting the next state using the ground truth action, $F(i_t, a_t)$, and predicting negative pairs chosen using noisy actions $F(i_t, \hat{a}^j)$.}
    \label{fig:train_overview}
    
\end{figure*}

Our paper achieves this by fine-tuning a pre-trained representation into: (a) a one-step dynamics module, $F(s, a)$, that predicts how the robot's next state given the current state/action; and (b) a ``functional distance module", $d(s,g)$, that calculates how close the robot is to achieving its goal $g$ in the state $s$. The distance function is learned with limited human demonstration data, using a contrastive learning objective (see Fig.~\ref{fig:train_overview}). Both $d$ and $F$ are used in conjunction to greedily plan robot actions (see Fig.~\ref{fig:teaser} for intuition). Our experiments demonstrate that this approach works better than policy learning (via Behavior Cloning), because the pre-trained representation itself does the heavy lifting (thanks to its structure) and we entirely dodge the challenge of multi-modal, sequential action prediction. Furthermore, our learned distance function is both stable and easy to train, which allows it to easily scale and generalize to new scenarios.

To summarize, we show a simple approach for exploiting the information hidden in pre-trained visual representations. Our contributions include:
\begin{itemize}
    \item Developing a simple algorithm for acquiring a distance function and dynamics model by fine-tuning a pre-trained visual representation on minimal (human collected) data.
    \item Creating an effective manipulation controller that substantially outperforms State-of-the-Art (SOTA) prior methods from the robot learning community (e.g. Behavior Cloning~\cite{ross2011reduction}, Offline-RL~\cite{kostrikov2021offline}, etc.).
    \item Demonstrating that our approach can handle four realistic manipulation tasks, generalize to new objects and settings, and solve challenging scenarios w/ multi-modal action distributions. 
\end{itemize}
\section{Related Work}
\label{sec:rw}

\paragraph{Behavior Cloning}

This paper adopts the Learning from Demonstration (LfD) problem setting~\cite{argall2009survey, billard2008survey, schaal1999imitation}, where a robot must acquire manipulation behaviors given expert demonstration trajectories. A standard approach in this space is to learn a policy ($\pi$) via Behavior Cloning~\cite{ross2011reduction} (BC), which directly optimizes $\pi$ to match the expert's action distribution. While conceptually simple, realistic action distributions are difficult to model, since they are inherently multi-modal and even small errors compound over time. Thus, policy learning requires extensive intensive network engineering (e.g. transformer architectures~\cite{dasari2020transformers,chen2021decision}, multi-modal prediction heads~\cite{shafiullah2022behavior,lynch2020learning}, etc.) and/or human-in-the-loop data collection algorithms~\cite{ross2011reduction,jang2022bc} to work in practice. Instead of learning a policy, we learn a \textit{functional distance metric} that captures the how ``close" a state is to reaching a target goal. This lets us build a manipulation controller using a simple greedy planner during test time, without any explicit action prediction! 

\paragraph{Offline RL}

Broadly speaking, the field of Reinforcement Learning~\cite{sutton2018reinforcement} (RL) seeks to learn a value/advantage/Q function by assuming access to a reward signal, which ``evaluates" a state (e.g. +1 reward for reaching goal). RL algorithms usually require environment interaction to learn, though the field of Offline-RL~\cite{levine2020offline} seeks to extend these systems to learn from offline trajectories. While these approaches have created some impressive robotics demos ranging from locomotion~\cite{lee2020learning,kumar2021rma} to dexterous manipulation~\cite{andrychowicz2020learning,nvidia2022dextreme,dasari2023pgdm}, RL methods are data hungry~\cite{rajeswaran2017learning,andrychowicz2020learning}, difficult to implement, require extreme ``reward engineering" (e.g. Meta-World~\cite{yu2019meta} reward functions are 50 lines!), hampered by unstable learning dynamics~\cite{kumar2020implicit}, and poorly generalize in realistic robotic manipulation settings~\cite{dasari2022rb2}. In contrast, our method learns a proxy for value functions (i.e. distances) purely from offline data, using representation learning algorithms (from the vision field) that are much more stable.

\paragraph{Learning Visual Rewards}
Our distance learning approach is analogous to learning visual rewards. Prior work learned success classifiers~\cite{sermanet2016unsupervised,xie2018few,torabi2018generative} and/or video similarity metrics~\cite{sermanet2018time,aytar2018playing,schmeckpeper2021reinforcement,bahl2022human,chen2021learning} from expert video demonstrations. Once learned, these modules were used to parameterize reward functions that could be used to train policies~\cite{abbeel2004apprenticeship}, and/or as cost functions for planners~\cite{angelov2020composing}. However, these papers mostly consider simple settings (e.g. simulation) or require the test setting to exactly match train time. This is because the learned reward functions are often noisy and/or poorly generalize to new scenes. In contrast, our learned distance metric is stable and can easily adapt to new scenarios, thanks to the pre-trained network's latent structure. This allows us to solve diverse tasks during test time on a real robot, using a simple (and fast) shooting method planner.

\paragraph{Distance Learning for Navigation}
Finally, prior work from Hahn et. al.~\cite{hahn2021nrns} explored distance learning to solve visual navigation tasks from offline video data. Our work is inspired from this approach, but makes three key innovations in order to apply it to (more challenging) robotic manipulation tasks. First, we simplify the network architecture to operate in representation space instead of using a graph network. Second, we learn the distances using a self-supervised contrastive loss, instead of relying on hand-defined objectives. And third, we add a dynamics function that proves to be critical for effective manipulation (see Table.~\ref{tab:ablation}
). 

\section{Methods}

\paragraph{Preliminaries} 

This paper considers learning goal-conditioned manipulation behaviors from image observations. The robot agent is provided a goal observation ($I_g$) -- e.g. target object in robot gripper -- and an observation ($I_t$) for the current time-step $t$. The robot must process these observations and decide an action to take ($a_t$). Note that all observations ($I$) are wrist-mounted RGB camera images with no depth or proprioceptive data. Additionally, the actions ($a$) are specified as arbitrary $SE(3)$ transforms for the robot's end-effector (more in Supplement ~\ref{supp:hardware}). Our goal is to learn a robotic manipulation controller, using a set of training trajectories $\mathcal{D} = \{ \tau_1, \dots, \tau_N \}$, where $\tau_i = \{I_g, I_1, a_1, \dots, a_{T-1}, I_T\}$. The test and train settings are visualized in Fig.~\ref{fig:prob_setting}.

\paragraph{Our Approach} 

Our method leverages a pre-trained representation network, $R$, to encode observations, $i_t = R(I_t)$, and enable control via distance learning. Specifically, we use contrastive representation learning methods~\cite{oord2018representation,mnih2013learning} to learn a distance metric, $d(i_j, i_k)$, within the pre-trained embedding space. The key idea is to use this distance metric to select which of the possible future state is closest to the goal state. But how do we predict possible future states? We explicitly learn a dynamics function, $F(i_t, a_t)$ that predicts future state for a possible action $a_t$. During test time, we predict multiple future states using different possible action and select the one which is closes to goal state. The following sections describe the learned dynamics module (see Sec.~\ref{sec:methods_dyn}), distance learning method (see Sec.~\ref{sec:methods_dist}), and test-time controller for robot deployment (see Sec.~\ref{sec:methods_test}) in detail. Fig.~\ref{fig:train_overview} provides a visual depiction of our training algorithm. Pseudo-code and hyper-parameters are presented in Supplement~\ref{supp:algorithm}.

\subsection{Dynamics Prediction}
\label{sec:methods_dyn}

An ideal dynamics function would perfectly capture how a robot's actions effects its environment. In our setting, this translates to predicting the next observation embedding, given the current embedding and commanded action: $F(i_t, a_t) = i_{t+1}$. Thus, $F$ can be learned via regression by minimizing reconstruction loss in embedding space: $\mathcal{L}_F = || F(i_t, a_t) - i_{t+1} ||_2$ (see Fig.~\ref{fig:train_overview}, left). This module has two primary uses: (1) during training it acts as a physically grounded regularizer that relays action information into the embedding space, and (2) during test time it enables the robot to plan actions in embedding space. As you will see, both properties are leveraged extensively in the rest of our method.

\subsection{Learning Task-Centric Distances}
\label{sec:methods_dist}

Our distance module $d(i_j, i_k)$ seeks to learn functional distances that encode task-centric reasoning (e.g. must reach for object before pushing it) alongside important physical priors (e.g. object is to the left so move left). How can we learn such a distance space? Since we have access to expert trajectories $\tau$, we know that the next state sequentially visited by the expert ($i_{t+1}$) is closer to the goal state ($i_g$) than arbitrary states ($\hat{i}$) reachable from $i_t$ -- i.e. $d(i_{t+1}, i_g) << d(\hat{i}, i_g)$. 

Our insight is that a distance metric with these properties can be learned via contrastive learning. Specifically, we define $d(i_j, i_k) = - cos(i_j, i_k)$, where $cos$ is cosine similarity, and sample a observation-state-goal tuples $(i_t, a_t, i_g)$, alongside random noise action candidates $\hat{a}^{1}, \dots, \hat{a}^{n}$ sampled from $\mathcal{D}$. We apply NCE loss and get: $$\mathcal{L}_d = \frac{exp( - d(F(i_t, a_t), i_g))}{exp( - d(F(i_t, a_t), i_g)) + \Sigma_j exp( - d(F(i_t, \hat{a}^{j}), i_g))}$$
This loss creates a warped embedding space where $i_{t+1}$ (hallucinated by $F(i_t, a_t)$) is pushed closer towards the goal state $i_g$, than other arbitrary states reachable from $i_t$ (again hallucinated by $F$). This process is shown visually in Fig.~\ref{fig:train_overview} (right), and precisely satisfies our criteria for good distance functions. 

\subsection{Training Details}
\label{sec:methods_train}
Both modules are trained jointly resulting in a final loss: $\mathcal{L} = \lambda_d \mathcal{L}_d + \lambda_F \mathcal{L}_F$. Note that $F$ is implemented as small neural networks with 2 hidden layer, while $d$ does not require any extra learned networks since its implemented via contrastive loss in the metric space. The shared representation network, $R$, we use is ResNet-18 initialized by R3M weights. The ADAM~\cite{kingma2014adam} stochastic gradient descent optimizer and back-propagation are used to train the network end-to-end. Please refer to Supplement~\ref{supp:training}. for additional details. 

\subsection{Test Time Robot Deployment}
\label{sec:methods_test}
During test time our learned modules must be able to solve real manipulation tasks when deployed on robot hardware. Thus, we develop a simple inference policy that solves for robot actions using our learned distance and dynamics function. First, a cost function $C_g(i) = d(i, i_g)$ is parameterized given a goal observation $I_g$ and the learned distance function. $C_g$ encodes how far an arbitrary image observation is from the goal image. The optimal action, $a^*_t$, will take the robot closest to the goal, hence: $a^*_t = \textbf{argmin}_a C_g(F(i_t, a))$. Note that we use the dynamics function $F$ to predict the next state, since we do not know the real $i^*_{t+1}$ during test time. The cost minimization is performed using the shooting method: random candidate actions, $a_1, \dots, a_N$, are sampled from $\mathcal{D}$, passed through the dynamics function, and the action with minimum cost $C_g(F(i_t, a_i))$ is executed on the robot. This process is illustrated in Fig.~\ref{fig:teaser}. Policy execution ends once the distance between the current state and the goal falls below a threshold value: $C_g(i_t) < \lambda$. While a more complicated planner could've been used, this solution was adopted due to its speed, simplicity, lack of hyper-parameters, and good empirical performance.

The final detail to discuss is gripper control for prehensile tasks (e.g. pick and place). Since the gripper status is binary (open/close) and highly correlated with the current state (i.e. close when object in hand), it makes more sense to handle it implicitly rather than as part of the action command $a_t$. Thus, we trained a gripper action classifier $\mathcal{G}(i_t) \in [0,1]$ that predicts the probability of closing the gripper given the current image embedding. This is also implemented by adding a single layer to ResNet-18 initialized by R3M weights and can be trained using a small amount ($< 100$) of human labelled images from the train dataset $\mathcal{D}$. As a result, our system can seamlessly handle gripper control in prehensile tasks, without requiring gripper labels for every frame in $\mathcal{D}$!

\begin{figure*}[t]
        \centering
        \begin{subfigure}[b]{0.24\linewidth}
            \includegraphics[width=\linewidth]{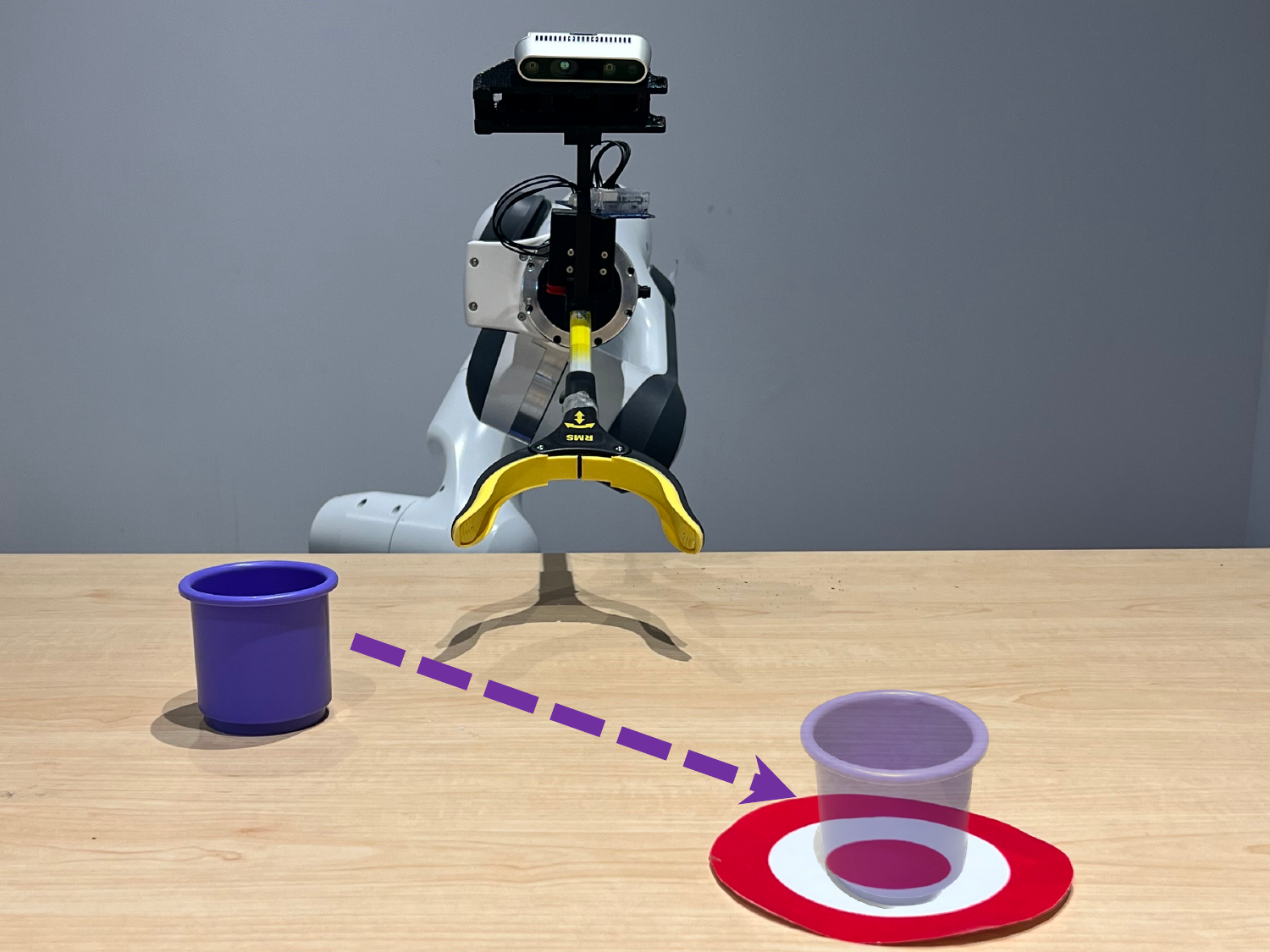}
            \vspace{-0.12in}
            \caption{\small Pushing}
            \label{fig:task_push}
        \end{subfigure}
        \begin{subfigure}[b]{0.24\linewidth}
            \includegraphics[width=\linewidth]{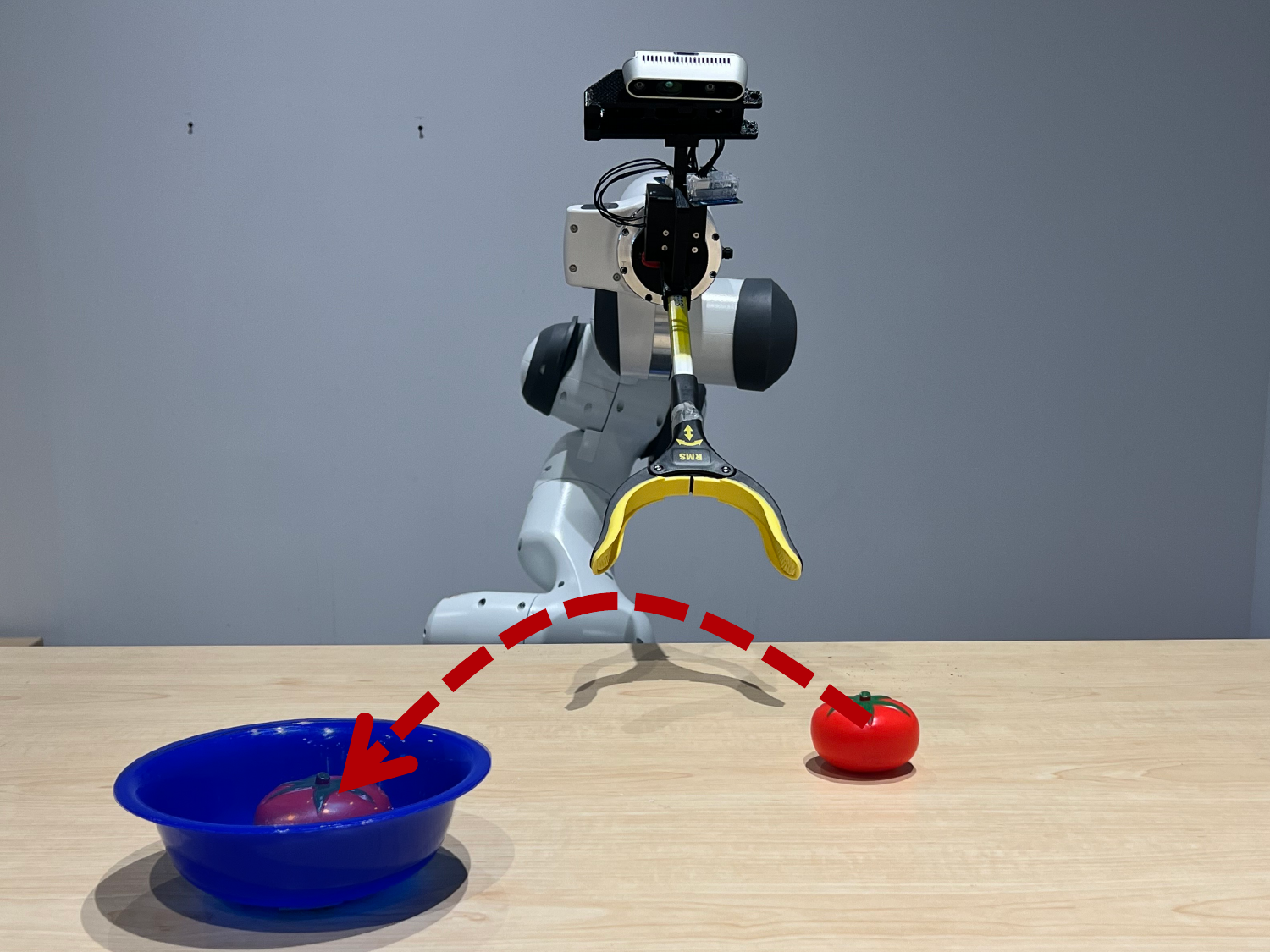}
            \vspace{-0.12in}
            \caption{\small Pick and Place}
            \label{fig:task_pick}
        \end{subfigure}
        \begin{subfigure}[b]{0.24\linewidth}
            \includegraphics[width=\linewidth]{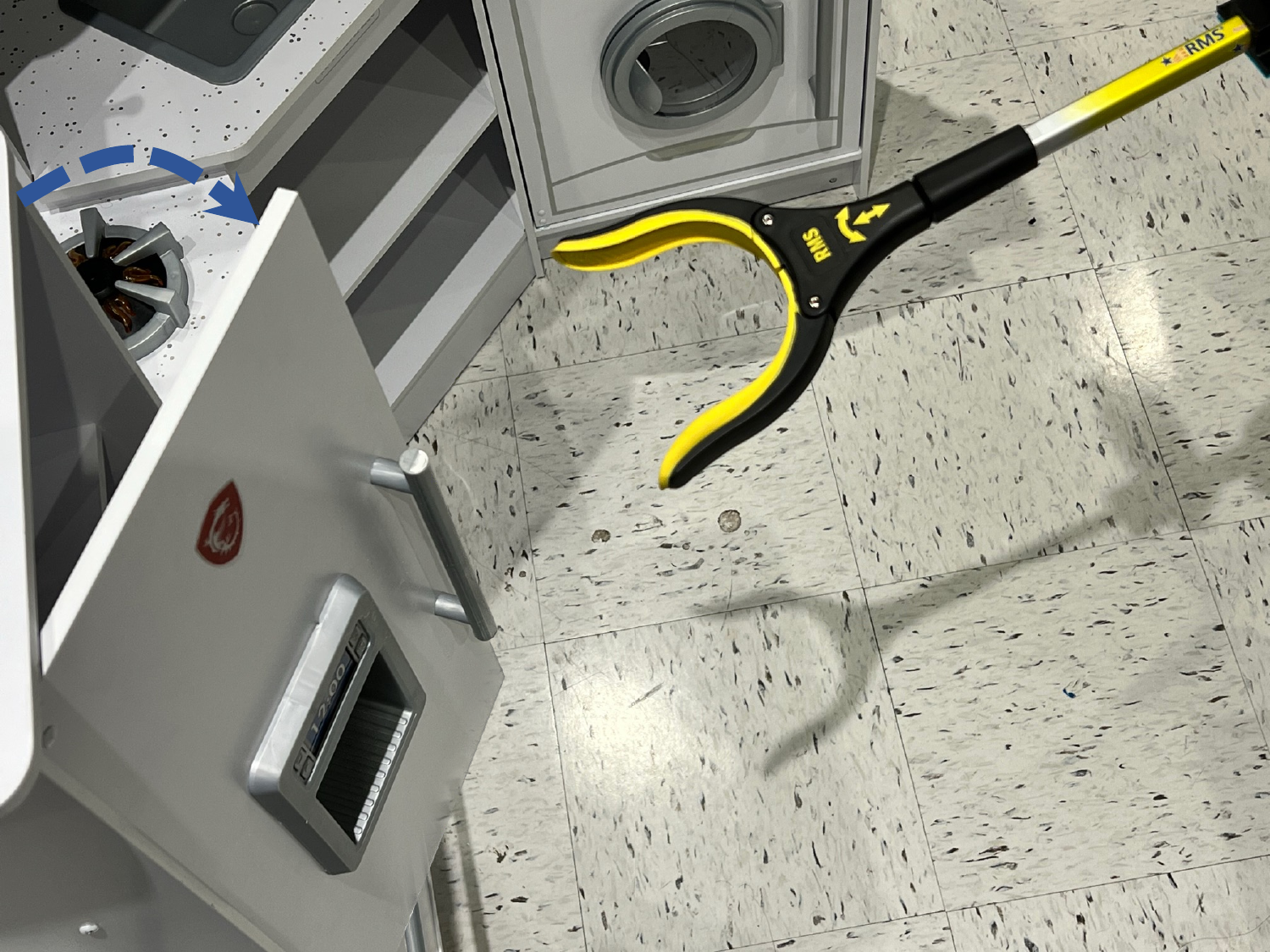}
            \vspace{-0.12in}
            \caption{\small Door Opening}
            \label{fig:task_open}
        \end{subfigure}
        \begin{subfigure}[b]{0.24\linewidth}
            \includegraphics[width=\linewidth]{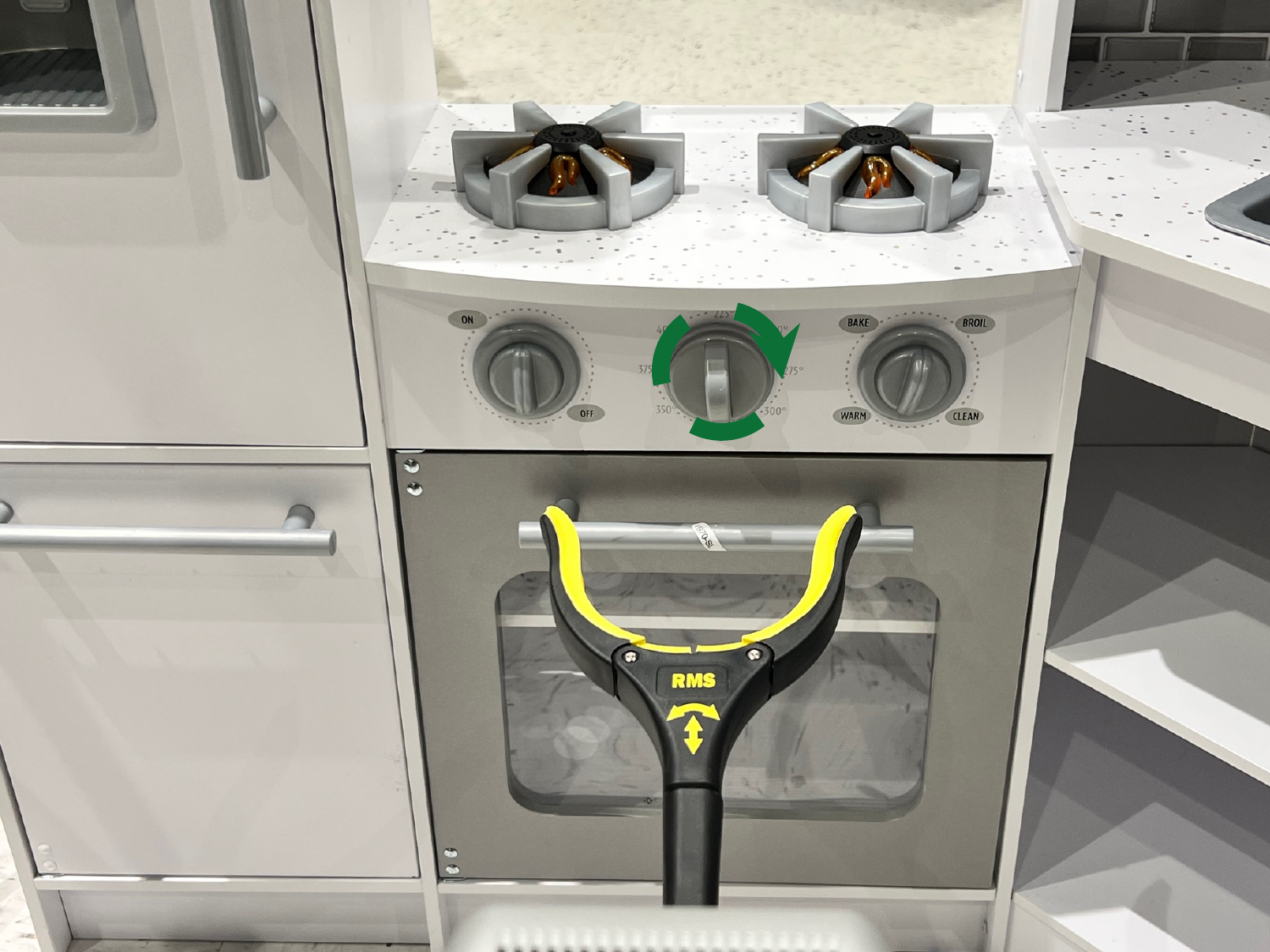}
            \vspace{-0.12in}
            \caption{\small Knob Turning}
            \label{fig:task_knob}
        \end{subfigure}
        \vspace{-0.05in}
        \caption{\small Our method is tested on 4 different manipulation tasks (pictured above). These tasks test different skill axis, ranging from task-level reasoning (e.g. detect target, move to object before goal, etc.) to fine-grained motor control (e.g. grab top of knob to turn).}
        \label{fig:tasks}
        
\end{figure*}

\begin{figure}[t]
    \centering 
    \includegraphics[width=\linewidth]{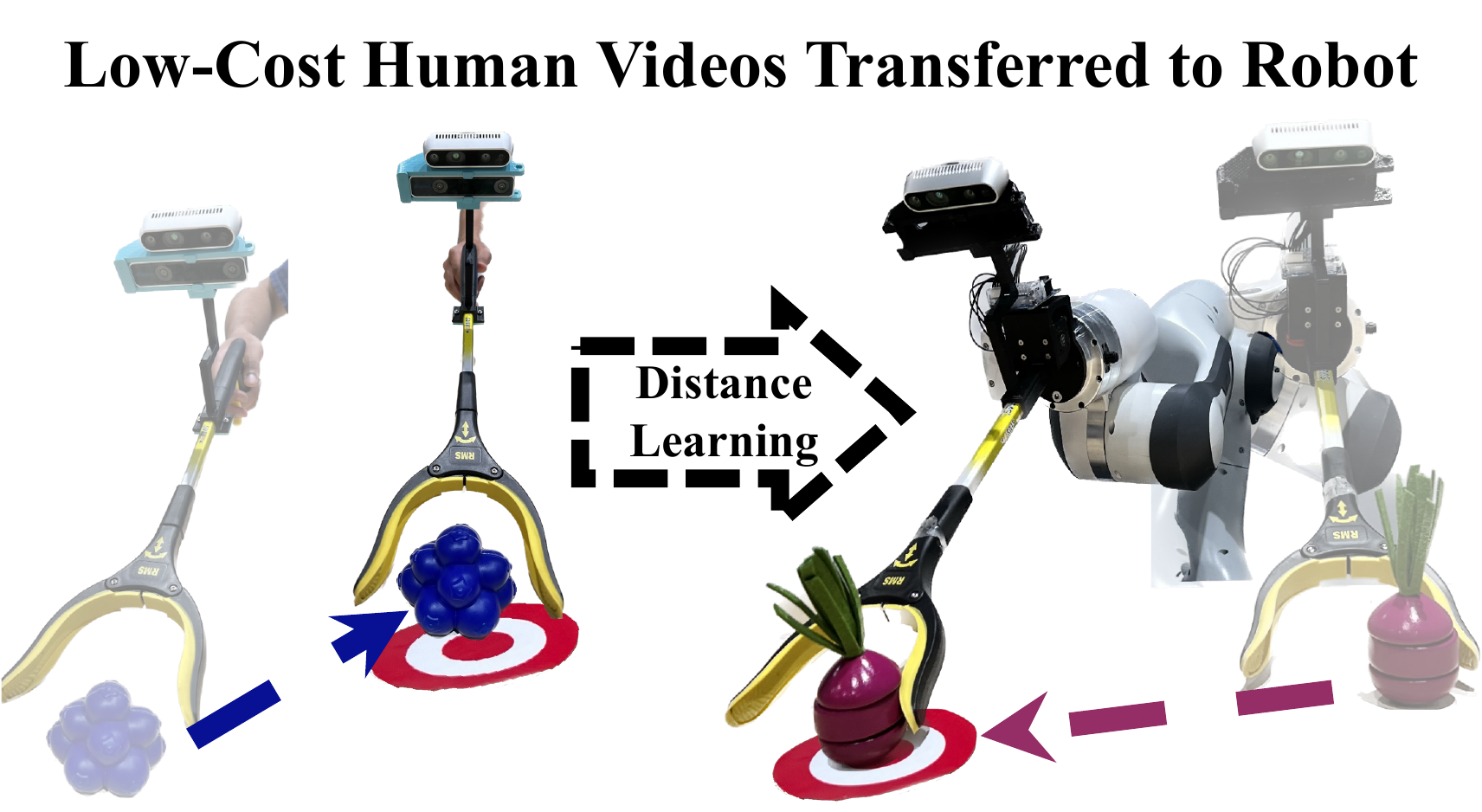} 
        \caption{In our problem setting we use a low-cost reacher grabber tool (left) to collect training demonstrations. These demonstrations are used to acquire a robot controller purely through distance/representation learning. The final system is deployed on a robot (right) to solve various tasks at test-time.}
    \label{fig:prob_setting}
    
\end{figure}

\begin{figure*}[t]
    \centering 
    \includegraphics[width=\linewidth]{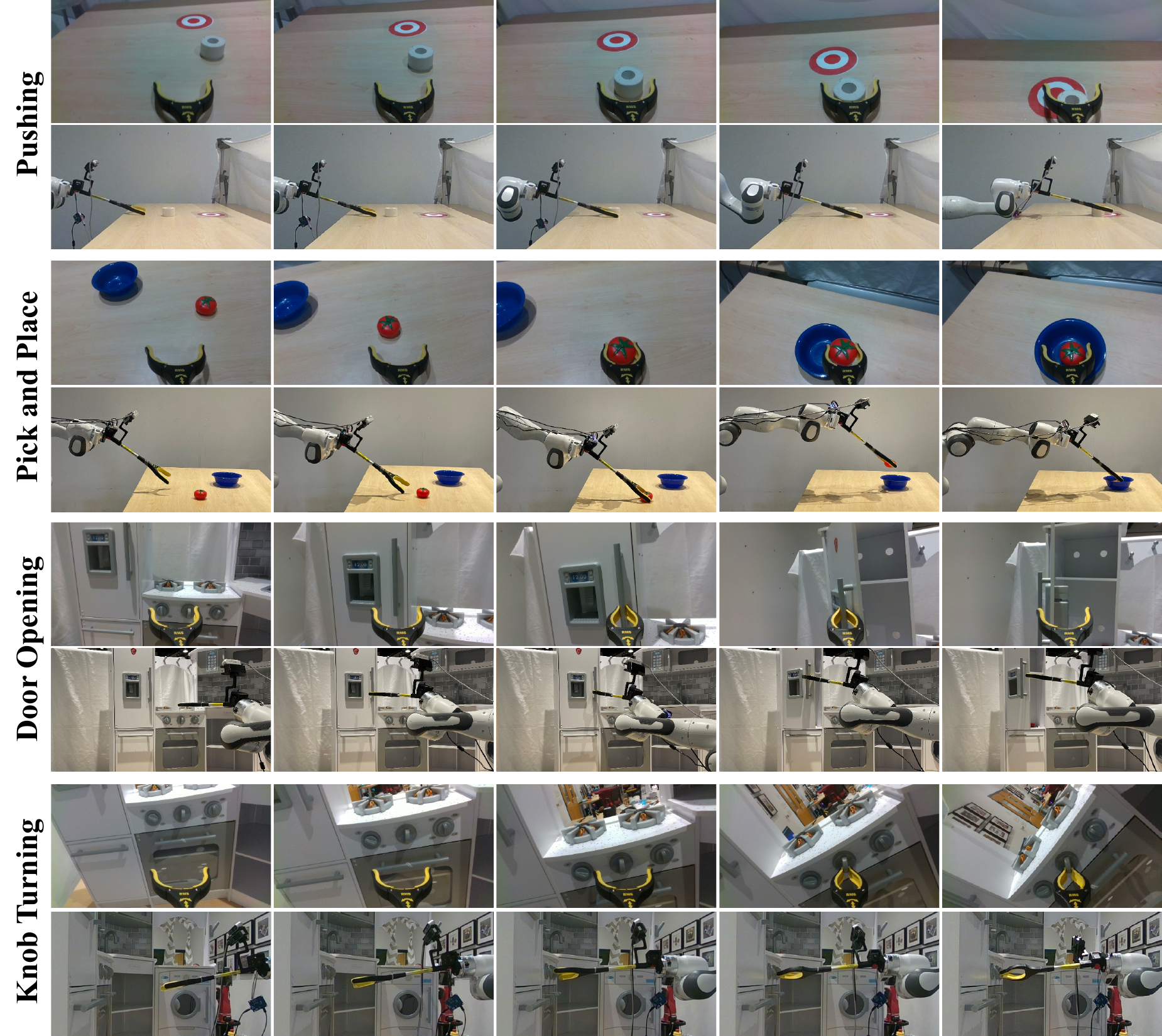} 
        \caption{Trajectories executed on the robot using our learned distance function. For each task, we show the $1^{st}$ person view (top) and $3^{rd}$ person view images (bottom). We show the learned visual embedding can encode functional distances between states for challenging tasks, like pushing, pick and place, door opening, and knob turning.}
    \label{fig:results}
    
\end{figure*}

\section{Experimental Setup}
\label{sec:exp_setup}

Our method is tested on four different manipulation tasks (see Fig.~\ref{fig:tasks}) that require a mix of \textit{high-level reasoning} (e.g. go to object before target), \textit{low level precision} (must carefully grab knob to turn it), and \textit{time-time generalization} (e.g. push novel object). 
We collect behavior data for each task from human demonstrators as described below. The trained policies are deployed on a real Franka Pandas robot (see Fig.~\ref{fig:prob_setting}, right). Additional details on the hardware setup and control stack are presented in Supplement~\ref{supp:hardware}.

\subsection{Tasks}
\label{sec:exp_setuptasks}
We now introduce the tasks used in our experiments (pictured in Fig.~\ref{fig:tasks}), and provide more details on their train and test conditions:

\paragraph{Pushing}

For this task (see Fig.~\ref{fig:task_push}) the robot must approach a novel object placed on the table and push it to the goal location (marked by target printout). \textbf{Training:} Our method was trained on a dataset of 100 demonstrations with diverse objects and randomized targets. \textbf{Testing:} During test time, the method was evaluated using 20 trails that involved unseen objects and new target printouts placed in randomly sampled positions. A trail is deemed successful if the robot pushes the object onto the target.  This is the only task that did not require gripper control. 

\paragraph{Pick and Place}

This task (see Fig.~\ref{fig:task_pick}) is analogous to the pushing task (described above), except the robot must now pick the object up from its initial location and place it in a target bowl. This task requires more precision than pushing, since grasping an object is harder than localizing it. \textbf{Training:} We collect a dataset of 400 demonstrations for training, using randomized train objects and target bowls. \textbf{Testing:} The method is evaluated using 20 trails that used novel objects and unseen target bowls (i.e. analogous to pushing). A trail is successful if the robot places an object into the target. 

\paragraph{Door Opening}

The opening task (see Fig.~\ref{fig:task_open}) requires the robot to approach a handle, grasp it, and then pull the door open. While conceptually simple, this task requires a great deal of precision, since the commanded actions must (almost) exactly match the direction of the door hinge. \textbf{Training:} We created a toy kitchen setting for this task, and collected 100 demonstrations of opening the door when starting from various initial positions. \textbf{Testing:} We evaluated on 20 real world test trials, where the robot's initial position and the door's initial state (e.g. position and hinge angle) were randomized. A trial was deemed successful if the door is opened fully.

\paragraph{Knob Turning}

In this final task (see Fig.~\ref{fig:task_knob}) the robot must approach a knob in the toy kitchen, and then turn it clock-wise. Similar to the opening task, knob turning requires very precise motor control, since the robot must grab the knob at exactly the right location and carefully rotate in order to turn. \textbf{Training:} We collect a dataset of 100 knob turning demonstrations. \textbf{Testing:} The method is evaluated on 20 trials, with randomized robot/knob initial positions. Success is determined by if the knob is turned far enough to cause an audible ``click."

\subsection{Robot-Free Data Collection}
\label{sec:problem_data}

A major strength of our setup is that training data ($\mathcal{D}$) can come from a different agent than the test time robot. We leverage this property to collect videos \textit{directly} from humans operating a low-cost grabber stick (see Fig.~\ref{fig:prob_setting}, left), and use structure from motion~\cite{schoenberger2016sfm} to recover actions. This allows us to collect demonstrations \textit{without} an expensive tele-op setup~\cite{zhang2018deep,kumar2015mujoco} or time-intensive kinesthetic demonstrations (i.e. manually moving the robot by hand). While based on prior work, our stick setup makes two important contributions: it is easier build and use than the highly instrumented setup in Song et. al.~\cite{song2020grasping}, and it provides higher quality images and action labels than the GoPro used in Young et. al.~\cite{young2021visual}. Please refer to Supplement~\ref{supp:data} for more information on our training and testing setups. We will release our dataset publicly.

\section{Experiments}
\label{sec:exp}

The following experiments seek to validate our distance learning framework on real control tasks (described above), and give some fundamental insights as to why they work. First, our baseline study (see Sec.~\ref{sec:exp_baseline}) compares our method against representative SOTA techniques in the field. In addition, in Sec.~\ref{sec:exp_multimodal} we analyze a specific scenario with highly multi-modal actions, where our model especially shines. Finally, the ablation study (see Sec.~\ref{sec:exp_ablation}) investigates which components of our system are actually needed for good performance.

\subsection{Baseline Study}
\label{sec:exp_baseline}

\begin{table}[!t]
        \centering
        \resizebox{\linewidth}{!}{%
            \begin{tabular}{l|cccc}
                \toprule
                 Task & \textbf{Ours} & BC~\cite{ross2011reduction,young2021visual} & IBC~\cite{florence2021implicit} & IQL~\cite{kostrikov2021offline}\\
                 \midrule
                 \textit{Pushing} & $\textbf{85}\%$ & $70\%$ & $70\%$ & $65\%$ \\
                 \textit{Pick and Place} & $\textbf{60}\%$ & $30\%$ & $30\%$ & $45\%$ \\
                 \textit{Door Opening} & $\textbf{55}\%$ & $40\%$ & $45\%$ & $\textbf{50}\%$ \\
                 \textit{Knob Turning} & $\textbf{40}\%$ & $20\%$ & $20\%$ & $\textbf{35}\%$ \\
                \bottomrule
            \end{tabular}
        }
        
        \caption{ We compare success rates for our method versus the baselines on all four manipulation tasks. Our distance learning method outperforms a suite of representative robot learning baselines. } 
        \label{tab:baseline}
\end{table}

\begin{table}[!t]
        \centering
        \small
        \resizebox{\linewidth}{!}{%
            \begin{tabular}{l|ccc|ccc}
                
                 & \multicolumn{3}{c|}{\textit{Pushing}} & \multicolumn{3}{c}{\textit{Pick and Place}} \\
                 
                 $\#$ Demos & \textbf{Ours} & BC & IBC & \textbf{Ours} & BC & IBC\\
                 
                 \midrule
                 50 & $\textbf{70}\%$ & $50\%$ & $40\%$ & $0\%$ & $0\%$ & $0\%$ \\
                 100 & $\textbf{85}\%$ & $70\%$ & $70\%$ & $\textbf{10}\%$ & $\textbf{10}\%$ & $0\%$ \\
                 200 & $90\%$ & $80\%$ & $\textbf{100}\%$ & $\textbf{20}\%$ & $\textbf{20}\%$ & $10\%$\\
                 400 & $\textbf{100}\%$ & $\textbf{100}\%$ & $\textbf{100}\%$ & $\textbf{60}\%$ & $30\%$ & $30\%$\\
                 600 & $\textbf{100}\%$ & $\textbf{100}\%$ & $\textbf{100}\%$ & $\textbf{70}\%$ & $50\%$ & $40\%$ \\
                \bottomrule
            \end{tabular}
        }

        \caption{ We compare our distance learning method versus the behavior cloning baselines (BC~\cite{ross2011reduction,young2021visual} and IBC~\cite{florence2021implicit}) when trained on varying amounts of data. Distance learning is able to learn faster than the baselines and scale better with data.}
        \label{tab:scaling_results}

\end{table}

We first note that our paper uses only demonstration data and no online data. Therefore, we compare our method against three SOTA baselines -- ranging from standard behavior cloning, to energy based modeling, and Offline Reinforcement Learning (RL) --  that cover a wide range of prior LfD work.  The baseline methods are described briefly below. To make the comparisons fair, we parameterize all neural networks with the same R3M representation backbone used by our method, and tune hyper-parameters for best possible performance. Please check Supplement~\ref{supp:baselines} for in depth details.

\begin{itemize}
    \item \textbf{Behavior Cloning~\cite{ross2011reduction,young2021visual} (BC):} BC learns a policy (via regression) that directly predicts actions from image observations: $\min_\pi ||\pi(I_t, I_g) - a_t||_2$. This provides a strong comparison point for a whole class of LfD methods that focus on learning motor policies directly.
    
    \item \textbf{Implicit Behavior Cloning~\cite{florence2021implicit} (IBC):} IBC learns an energy based model that can predict actions during test time via optimization: $a_t = \textit{argmin}_a E(a, I_t, I_g)$. This method is conceptually very similar to behavior cloning, but has the potential to better handle multi-modal action distributions and discontinuous actions. 
    
    \item \textbf{Implicit Q-Learning~\cite{kostrikov2021offline} (IQL):} IQL is an offline-RL baseline that learns a Q function $Q(s,a) = Q((I_t, I_g), a_t)$, alongside a policy that maximizes it $\pi(I_t, I_g) = \textit{argmax}_a Q(s, a)$. Note that IQL's training process require us to annotate our offline trajectories $\mathcal{D}$ with a reward signal $r_t$ for each time-step. While this can be done automatically in our setting (Supplement~\ref{supp:baselines}), this is an additional assumption that could be very burdensome in the general case. 
\end{itemize}

Our method is compared against the baselines on all four of our tasks (see Sec.~\ref{sec:exp_setuptasks}). Results are reported in Table~\ref{tab:baseline}. Note how our distance learning approach achieves the highest success rates on all of four tasks. In addition, our method is much simpler than the strongest baseline (IQL), which requires an expert defined reward signal and uses a more complicated learning algorithm. The final test time performance is visualized in Fig.~\ref{fig:results} and on our website: \website.

\paragraph{Data Scaling} A common cited strength of BC (versus RL) is its ability to continuously improve with additional expert data. Can our method do the same? Table~\ref{tab:scaling_results} compares our distance learning method versus the BC/IBC baselines (on pushing/pick-place tasks), when trained on various amounts of data. Note how our method both learns faster than BC/IBC, and continuously scales with added demonstrations. This suggests that distance learning is a powerful alternative to BC on robotic manipulation tasks, no matter how much training data is available.

\subsection{Multi-Modality Experiment}
\label{sec:exp_multimodal}

\begin{figure}[t]
    \centering
    \includegraphics[width=0.8\linewidth]{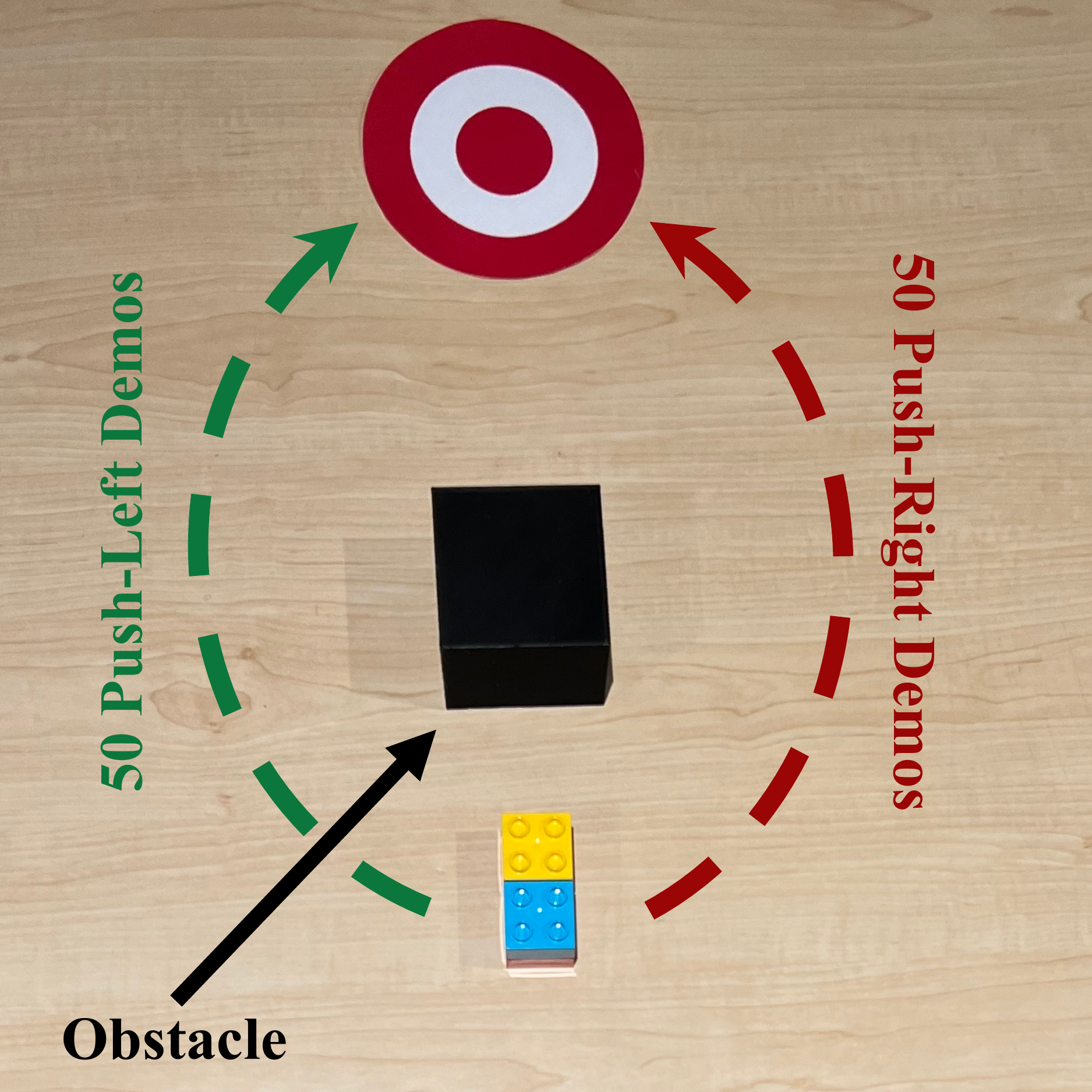}
    \caption{ Our method can solve tasks with highly multi-modal action distributions that are difficult for the baselines. In this example, our distance learning controller successfully pushes the block around the obstacle, while Behavior Cloning learns to incorrectly push the block forward (i.e. predicts mean action).}
    \label{fig:multimodal_task}
    
\end{figure}

A common challenge in policy learning is effectively handling multi-modal action predictions. We propose a simple test scenario that allows us to test our method in this setting without any other confounding variables. Specifically, we create a ``obstacle pushing" task where the robot must push an object to the target without disturbing an obstacle placed in the middle (see Fig.~\ref{fig:multimodal_task}). Observe that there are two equally valid ways that the robot could push the block; to the left or to the right. We collect a training dataset of 100 demos, equally split between going left and right. This creates action multi-modality where the agent must arbitrarily choose between two scenarios in order to successfully perform the task, but will fail if it simply averages the actions and pushes the block forward (into the obstacle). We expect that standard BC will fall into this failure mode, since it is trained using standard L2 action regression. In contrast, our method should still work since the low distance states will move the robot either to the left or right state, thus allowing it to  make a discrete choice.

To verify this hypothesis, we test our method, the BC baseline, and IQL (the strongest non-BC baseline) on 20 trials in this task setting. Success is judged similarly as to pushing (see Sec.~\ref{sec:exp_setuptasks}), with the added condition that the robot should not push the obstacles into the target. Our method achieves a $\textbf{95}\%$ success rate, versus $0\%$ for BC and $90\%$ for IQL in this setting! As expected, our method is able to greatly improve upon BC in this setting, because it explicitly avoids action prediction. While IQL should not suffer as badly from multi-modality, thanks to its learned value function, our method is still able to outperform it despite being simpler algorithmically and trained without reward signal.

\subsection{Ablation Study}
\label{sec:exp_ablation}

Our final experiment removes components of our method in order to determine which parts were most crucial for control performance. The \textit{representation ablation} replaces the R3M backbone in our neural network with a pre-trained ImageNet representation (using same ResNet-18 architecture for both). We expect the ImageNet weights to be be weaker, since they were trained on a smaller dataset that did not include ego-centric videos (ImageNet vs Ego4D used by R3M). Additionally, the \textit{dynamics ablation} removes the dynamics module during training, thus leaving the distance model/embeddings without any physical grounding. This is similar to the method from Hahn et. al.~\cite{hahn2021nrns} (see Sec.~\ref{sec:rw}), with an added dynamics module. We expect this ablation to perform worse on manipulation tasks, because dynamics is more important in this setting v.s. visual navigation.

The two ablations are evaluated and compared against our method on all four test tasks (see Table~\ref{tab:ablation}). We find that our method strongly outperforms both of our baselines, which suggests that our intuition outline above is correct. In particular, we conclude that stronger representations can directly stronger control performance using our method. Additionally, this shows that the dynamics grounding is important to ensure that our learned distances transfer onto real robot hardware during test time, especially in manipulation tasks that involve object interactions/contact.

\begin{table}[!t]
        \centering
       
        \resizebox{\linewidth}{!}{%
            \begin{tabular}{l|ccc}
                \toprule
                 & \textbf{Ours} & ImageNet Repr. & No Dynamics~\cite{hahn2021nrns}\\
                 \midrule
                  \textit{Pushing} & $\textbf{90}\%$ & $40\%$ & $60\%$ \\
                 \textit{Pick and Place} & $\textbf{60}\%$ & $40\%$ & $35\%$ \\
                 \textit{Door Opening} & $\textbf{55}\%$ & $45\%$ & $\textbf{50}\%$ \\
                 \textit{Knob Turning} & $\textbf{40}\%$ & $15\%$ & $\textbf{40}\%$ \\
                \bottomrule
            \end{tabular}
        }
        
        \caption{ This table compares success rates for our method versus the ablations on the all four tasks. As you can see, removing the dynamics module from our method during training and training with a weaker representation both result in worse performance.}
        \label{tab:ablation}
        
\end{table}
\section{Discussion}
\label{sec:discussion}

This paper demonstrates that neural image representations can be more than just state representations: a simple metric defined within the embedding space can help infer robot actions. We leverage this insight to learn a distance function and dynamics function using minimal low-cost human data. These modules parameterize a robotic planner that is validated across 4 representative manipulation tasks. Despite its simplicity, our method is able to outperform standard imitation learning and offline-RL methods in the robot learning field. This is especially true in situations involving multi-modal action distributions, where our method entirely outclasses a standard BC baseline. Finally, the ablation study demonstrates that stronger representations directly result in stronger control performance, and that dynamics grounding is important for our scheme to work in practice. 

\paragraph{Future Work}

We hope that this paper inspires future work in the space of representation learning and robotics. Follow up works should improve visual representations specifically for robotics, by more precisely capturing fine-grained interactions between gripper/hand and objects. This could improve performance on tasks like knob turning, where the pre-trained R3M encoder often struggled to detect small shifts of the gripper relative to the knob. We also hope that our contrastive learning method could be extended to learn entirely without action labels. This would allow us to train on large scale manipulation datasets (e.g. YouTube videos) and transfer the results directly to the control setting. Finally, it would be ideal if data collected using our low-cost stick (see Fig.~\ref{fig:prob_setting}) could be used with a stronger, more reliable (commercial) gripper, despite the domain gap. This would enable us to solve more tasks during test time.

{\small
\bibliographystyle{ieee_fullname}
\bibliography{references}
}

\clearpage
\appendix
\section{Method Details}

In this section, we present more details of our method.

\subsection{Algorithm}

Algorithm~\ref{alg:train} provides the psuedo-code of training our method. Algorithm~\ref{alg:test} provides the psuedo-code of deploying our method on real robot during test time.

\subsection{Data Collection Details}
\label{supp:data}

Our Robot-Free Data Collection is built around the 19-inch RMS Handi Grip Reacher and Intel RealSense D435 camera to collect visual data. We attach a 3D printed mount above the stick to hold the camera in place. At the base of the reacher-grabber, there is a lever to control the opening and closing of the gripper fingers. To collect demonstrations, a human user uses the setup shown in Fig~\ref{fig:prob_setting} (a) which allows the user to easily push, grab and interact with everyday objects in an intuitive manner. We also use an Intel RealSense T265 camera to track the end-effector position via visual inertial odometry. The demonstrations are represented as a sequence of images $I_t$ with corresponding end-effector positions $P_t$. 

Once we have the end-effector pose $P_t$ for every image $I_t$,we extract the relative transformation $T_{t,t+1} = P_t^{-1} \times P_{t+1}$ between consecutive frames and use them as the action for training. 

\subsection{Training Details}
\label{supp:training}

Our method is composed of two modules: 1) a pre-trained representation network, $R$, to encode observations, $i_t = R(I_t)$, and enables control via distance learning. 2) a dynamics function $F$, to predict the future state for a possible action $a_t$.

Here we encode the image $I_t$ via a ResNet18~\cite{he2016deep} and use a 1-layer projection head to get the visual embedding $i_t \in R^{128}$. For the dynamics function $F(i_t, a_t)$, we use a 3-layer MLP (128 + action dimension to 128 to 128) with ReLU activation. Both modules are trained jointly with $\mathcal{L} = \lambda_d \mathcal{L}_d + \lambda_F \mathcal{L}_F$ where $\lambda_d = \lambda_F = 1$. We use the Adam optimizer~\cite{kingma2014adam} for training the network with a batch size of 64 and a learning rate of $10^{-3}$. We train the network for 500 epochs and report the performance.

For each current observation, we randomly sample 4096 actions from training set as negative logits and use the ground truth action as possitive logit. For rotation-free tasks, like pushing and stacking, we use only the translation of $T_{t,t+1}$ as the action, such that $a_t \in R^3$. For rotation-heavy tasks, like knob turning, we use both translation and rotation of $T_{t,t+1}$ as the action, such that $a_t \in R^{12}$ (first three rows in the $SE(3)$ homogeneous transformation).

To improve the performance of our networks with limited data, we use the following data augmentations in training: (a) color jittering: randomly adds up to $\pm 20\%$ random noise to the brightness, contrast and saturation of each observation. (b) gray scale: we randomly convert image to grayscale with a probability of 0.05. (c) crop: images are randomly cropped to $224 \times 224$ from an original image of size $240 \times 240$. 

\label{supp:algorithm}

\begin{algorithm}[!t]
\caption{LMLS (Train of Passive Videos)}
\label{alg:train}
\algcomment{
}
\definecolor{codeblue}{rgb}{0.25,0.5,0.5}
\lstset{
  backgroundcolor=\color{white},
  basicstyle=\fontsize{7.2pt}{7.2pt}\ttfamily\selectfont,
  columns=fullflexible,
  breaklines=true,
  captionpos=b,
  commentstyle=\fontsize{7.2pt}{7.2pt}\color{codeblue},
  keywordstyle=\fontsize{7.2pt}{7.2pt},
}
\begin{lstlisting}[language=python]
###################Initialize###################
R: observation encoder; F: dynamics function
G: gripper action classifier
#####################Input######################
I_t: current images; I_t+1: current images; 
I_g: goal images; a_t: current actions; 
a_r: sampled random actions; g_t: gripper action
################################################
# Learning Task-Centric Distances
for x in loader: # load a minibatch x with N samples
    i_t = R(x.I_t) # encode current images
    i_t+1 = R(x.I_t+1) # encode next images
    i_g = R(x.I_g) # encode goal images
    i_p = F(i_t, x.a_t) # predict next state
    i_h = F(i_t, x.a_r) # hallucinated next state
    l_pos = cosine_similarity(i_p, i_g)#positive: N*1
    l_neg = cosine_similarity(i_h, i_g)#negative: N*M
    logits = cat([l_pos, l_neg], dim=1)#logits: Nx(1+K) 
    labels = zeros(N) # contrastive loss
    loss_dis = CrossEntropyLoss(logits, labels)
    loss_dyn = MSELoss(i_p, i_t+1) # dynamics loss
    loss = loss_dis + loss_dyn
    loss.backward() 
    update(R.params, F.params) # Adam update
# Learning Binary Gripper Classifier
for x in loader: # load a minibatch x with N samples
    i_t = R(x.I_t) # encode current images
    g = G(i_t) # predict gripper action
    loss = BCELoss(g, x.g_t)
    loss.backward() # Adam update
    update(R.params, F.params)
    

\end{lstlisting}

\end{algorithm}

\begin{algorithm}[!t]
\caption{LMLS (Test on Robots)}
\label{alg:test}
\algcomment{
}
\definecolor{codeblue}{rgb}{0.25,0.5,0.5}
\lstset{
  backgroundcolor=\color{white},
  basicstyle=\fontsize{7.2pt}{7.2pt}\ttfamily\selectfont,
  columns=fullflexible,
  breaklines=true,
  captionpos=b,
  commentstyle=\fontsize{7.2pt}{7.2pt}\color{codeblue},
  keywordstyle=\fontsize{7.2pt}{7.2pt},
}
\begin{lstlisting}[language=python]
###################Initialize###################
T_0: robot home position; I_0: initial observation
#####################Input######################
R: observation encoder; F: dynamics function
G: gripper action classifier; I_g: goal;
a_r: sampled random actions
################################################
i_g = R(I_g)
While not reach_goal or t < max_step:
    i_h = F(R(I_t), a_r) # hallucinated next state
    distance = - cosine_similarity(i_h, i_g)
    # choose action leads to smallest distance-to-goal
    best_action_index = argmin(distance)
    a_t = a_r[best_action_index]
    g = G(I_t) # predict gripper action
    # Send command to robot and get new observation
    T_t+1, I_t+1 = Robot(T_t, a_t, g) 
    
\end{lstlisting}

\end{algorithm}

\section{Experiment Details}
\label{supp:experiment}

\subsection{Hardware Setup and Control Stack}
\label{supp:hardware}

Our real-world experiments make use of a Franka Panda robot arm with all state logged at 50 Hz. Observations are recorded from an Intel RealSense D435 camera, using RGB-only images at $1280\times720$ resolution, logged at 30 Hz. On the robot’s end, we use the same 19-inch RMS Handi Grip Reacher and attach it using metal studs to the robot end effector through a 3D-printed mount. To control the fingers of the tool, we remove the lever at the base of the grip reacher and replace it with a dynamixel XM430-W350-R servo motor. 

The learned visual-feedback policy operates at 5 Hz. On a GTX 1080 Ti GPU. The learned action space is a 6 Dof homogeneous transformation, from the previous end-effector pose to the new one. We then calculate the new joint position using inverse kinematics through Mujoco~\cite{todorov2012mujoco}. The joint positions are linearly interpolated from their 5 Hz rate to be 100 Hz setpoints to our joint level controller. The joint positions are then sent to Facebook Polymetis~\cite{polymetis} to control the Franka robot.

\begin{figure}[t]
    \centering 
    \includegraphics[width=\linewidth]{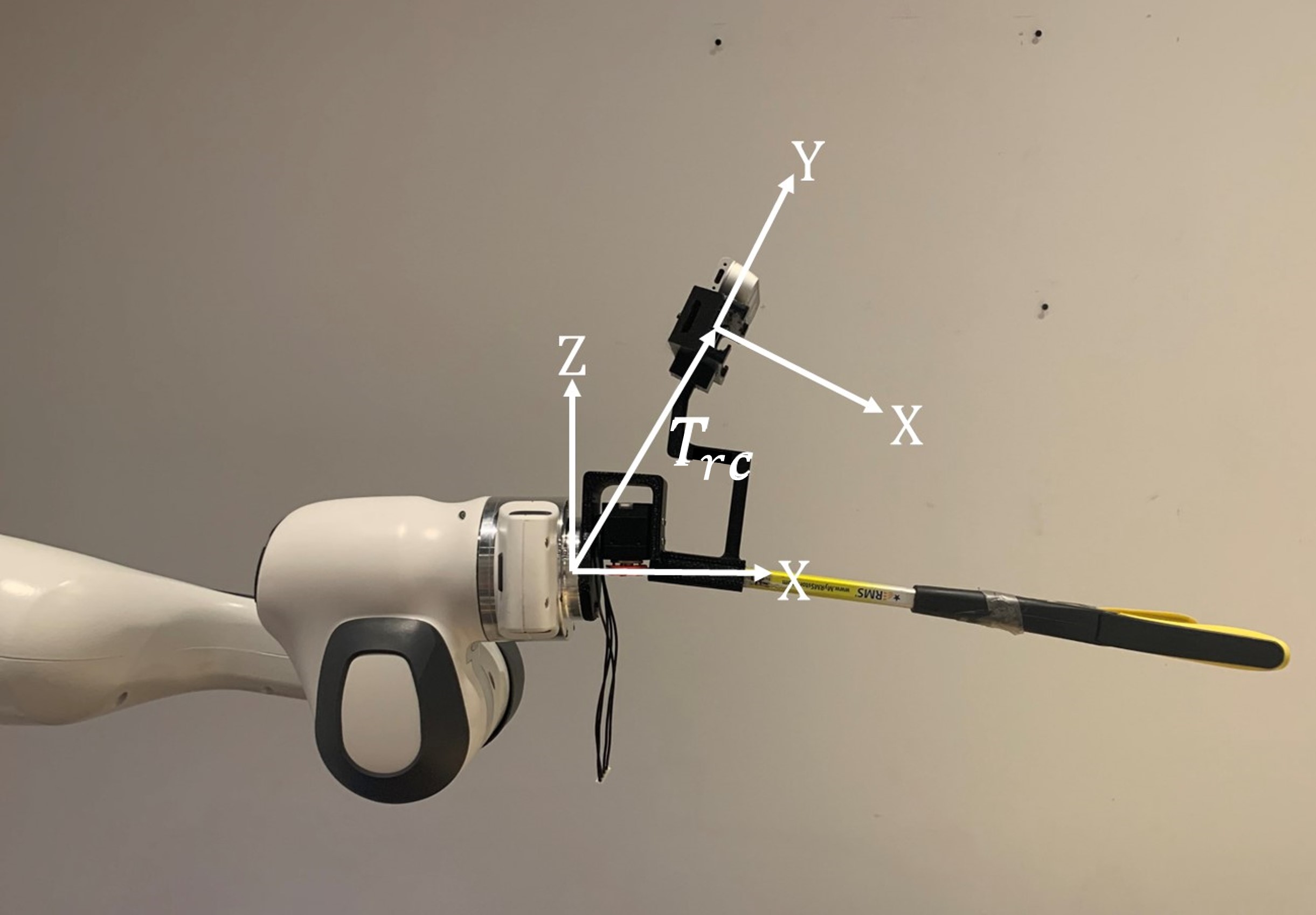} 
        \caption{Transform actions in camera frame to robot frame.}
    \label{fig:transformation}
\end{figure}

It worth noticing the learned action space is in camera frame instead of robot frame. Thus, we need to transform the predicted actions $T_{c^0c^1}$ to robot frame through a fixed homogeneous transformation $T_{cr}$ (Fig~\ref{fig:transformation}).

Using chain rule, we can easily calculate the motion in robot frame ($T_{r^0r^1}$) as:

\begin{align}
    T_{r^0r^1} &=  T_{rc} \times T_{c^0c^1} \times T_{cr}\\
    &=T^{-1}_{cr} \times T_{c^0c^1} \times T_{cr}
\end{align}

\subsection{Baselines}
\label{supp:baselines}

We compare our method against three SOTA baselines: behavior cloning, implicit behavior cloning, implicit Q-learning. To make the comparisons fair, we parameterize all neural networks with the same R3M representation backbone used by our method, and tune hyper-parameters for best possible performance. 

\begin{itemize}
    \item \textbf{Behavior Cloning~\cite{ross2011reduction,young2021visual} (BC):} BC learns a policy (via regression) that directly predicts actions from image observations: $\min_\pi ||\pi(I_t, I_g) - a_t||_2$. This provides a strong comparison point for a whole class of LfD methods that focus on learning motor policies directly (i.e. learn policies that predict actions). Here we encode the image $I_t$ via a ResNet18~\cite{he2016deep} and use a 4-layer multi-layer perceptron~\cite{rosenblatt1961principles} to regress the actions (512-256-128-action dimension). The predicted actions are supervised with ground-truth actions via MSELoss. We use the Adam optimizer~\cite{kingma2014adam} for training the network with a batch size of 64 and a learning rate of $10^{-3}$. We train the network for 200 epochs and report the performance.
    
    \item \textbf{Implicit Behavior Cloning~\cite{florence2021implicit} (IBC):} IBC learns an energy based model that can predict actions during test time via optimization: $a_t = \textit{argmin}_a E(a, I_t)$. This method is conceptually very similar to behavior cloning, but has the potential to better handle multi-modal action distributions and discontinuous actions. Similarly, we encode the image $I_t$ via a ResNet18~\cite{he2016deep} and use a 1-layer projection head to get the visual embedding $i_t \in R^{128}$. We also encode the actions with a 3-layer multi-layer perceptron (action dimension to 32 to 64 to 128). For each current observation, we randomly sample 4096 actions $\hat{a}_j$ from training set as negative logits and use the ground truth action $a_t$ as possitive logit. Both visual encoder and action encoder are trained with NCE loss:
    
    $$\mathcal{L} = \frac{exp(cos(i_t, a_t))}{exp(cos(i_t, a_t)) + \Sigma_j exp(cos(i_t, \hat{a}_j))}$$
    
    We use the Adam optimizer~\cite{kingma2014adam} for training the network with a batch size of 64 and a learning rate of $10^{-3}$. We train the network for 500 epochs and report the performance.
    
    \item \textbf{Implicit Q-Learning~\cite{kostrikov2021offline} (IQL):} IQL is an offline-RL baseline that learns a Q function $Q(s,a) = Q((I_t, I_g), a_t)$, alongside a policy that maximizes it $\pi(I_t, I_g) = \textit{argmax}_a Q(s, a)$. Note that IQL's training process require us to annotate our offline trajectories $\mathcal{D}$ with a reward signal $r_t$ for each time-step. Here we label the trajectories with sparse reward: +1 for end-effector reaching the target object, +2 for reaching the goal state, and +0 for all other states. We use d3rlpy~\cite{seno2021d3rlpy} and trained the model for 500k steps.

\end{itemize}

\end{document}